\documentclass{article}
\usepackage{iclr2026_conference,times}

\usepackage{amsmath,amsfonts,bm}

\def\eqref#1{equation~\ref{#1}}

\def\1{\bm{1}}

\DeclareMathAlphabet{\mathsfit}{\encodingdefault}{\sfdefault}{m}{sl}
\SetMathAlphabet{\mathsfit}{bold}{\encodingdefault}{\sfdefault}{bx}{n}

\def\gG{{\mathcal{G}}}

\def\gT{{\mathcal{T}}}

\def\gX{{\mathcal{X}}}
\def\gY{{\mathcal{Y}}}

\newcommand{\E}{\mathbb{E}}

\newcommand{\R}{\mathbb{R}}

\DeclareMathOperator*{\argmax}{arg\,max}

\usepackage{hyperref}
\usepackage{url}
\usepackage{xurl}
\usepackage{algorithm}
\usepackage{algorithmic}
\usepackage{booktabs}
\usepackage{multirow}
\usepackage{graphicx}
\usepackage{subcaption}
\usepackage{wrapfig}
\usepackage{framed}
\title{Learning to Self-Evolve}

\author{Xiaoyin Chen\textsuperscript{1, 2}\thanks{This work was done during Xiaoyin's internship at Snowflake. Correspondence should be addressed to Canwen Xu: \texttt{canwen.xu@snowflake.com}.} \quad
Canwen Xu\textsuperscript{3} \quad
Yite Wang\textsuperscript{3} \quad
Boyi Liu\textsuperscript{3} \quad
Zhewei Yao\textsuperscript{3} \quad
Yuxiong He\textsuperscript{3} \\
\textsuperscript{1}Mila -- Quebec AI Institute \quad
\textsuperscript{2}University of Montreal \quad
\textsuperscript{3}Snowflake
}
\iclrfinalcopy
\begin{document}

\maketitle

\begin{abstract}
We introduce Learning to Self-Evolve (LSE), a reinforcement learning framework that trains large language models (LLMs) to improve their own contexts at test time. We situate LSE in the setting of test-time self-evolution, where a model iteratively refines its context from feedback on seen problems to perform better on new ones. Existing approaches rely entirely on the inherent reasoning ability of the model and never explicitly train it for this task. LSE reduces the multi-step evolution problem to a single-step RL objective, where each context edit is rewarded by the improvement in downstream performance. We pair this objective with a tree-guided evolution loop. On Text-to-SQL generation (BIRD) and general question answering (MMLU-Redux), a 4B-parameter model trained with LSE outperforms self-evolving policies powered by GPT-5 and Claude Sonnet 4.5, as well as prompt optimization methods including GEPA and TextGrad, and transfers to guide other models without additional training. Our results highlight the effectiveness of treating self-evolution as a learnable skill.\footnote{Code is available at \url{https://github.com/chenyn66/learning-to-self-evolve}.}
\end{abstract}

\section{Introduction}

The ability to adapt and evolve in response to environmental feedback has long been considered central to human intelligence~\citep{piaget1952origins, jintelligence7040023}. A chess player improves by analyzing past games; a software engineer grows more proficient with a codebase through months of daily work. In both cases, experience accumulates and the person adjusts their approach accordingly. Current large language model (LLM) training pipelines exhibit a similar dynamic, particularly at the post-training stage, where reinforcement learning (RL) refines the behavior of the model on its own generated data~\citep{lightman2023let, o1, deepseek2025r1}. However, this learning stops once training ends. At deployment, an LLM applies the same policy regardless of how many problems it has solved in a domain, and discards all accumulated experience once the context resets. This gap between static deployment and dynamic adaptation motivates the study of \emph{test-time self-evolving} systems: systems that continuously update themselves in response to new observations at test time.
\begin{figure}[t]
  \centering
  \includegraphics[width=\linewidth, trim=0 55 0 0, clip]{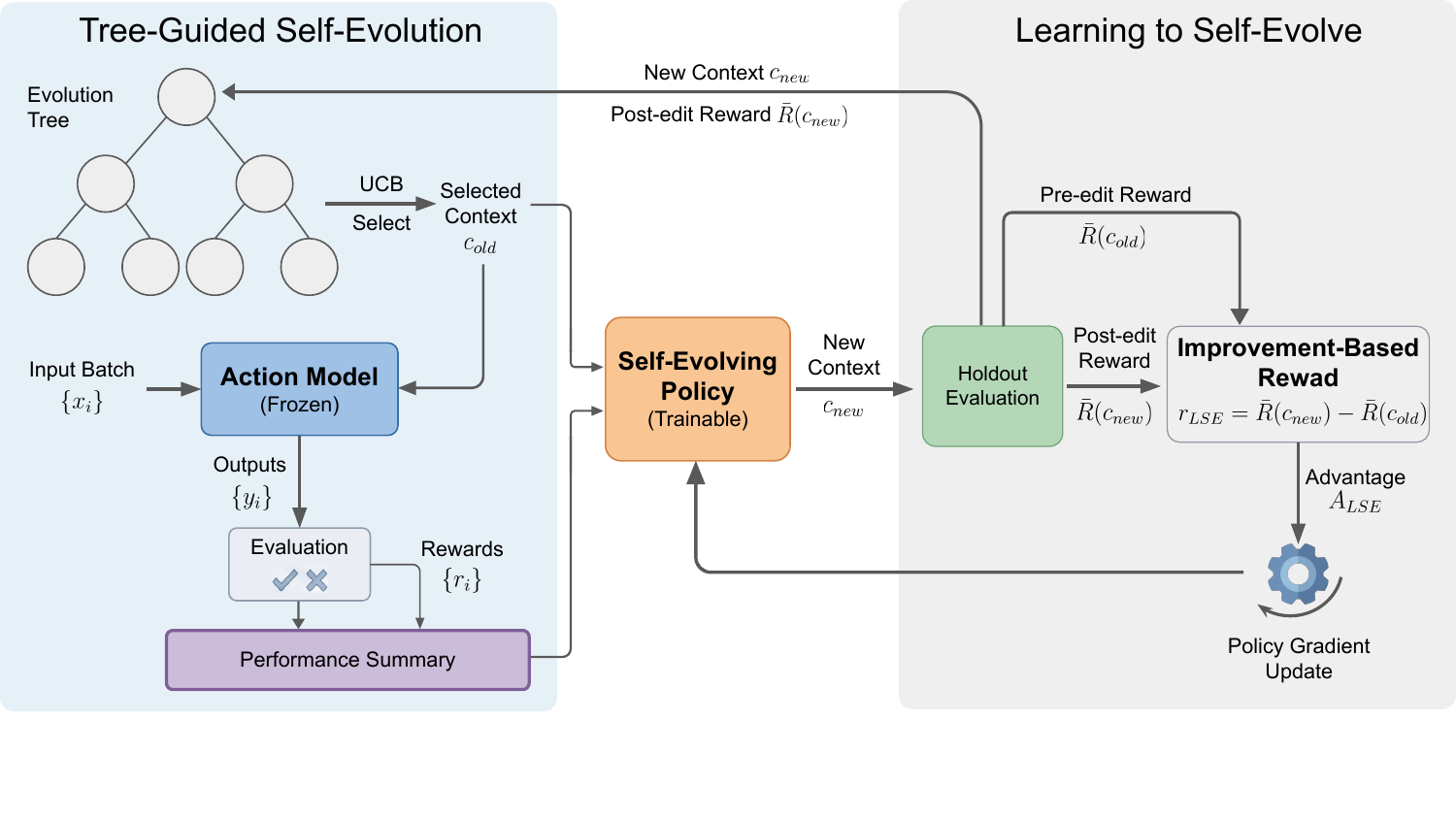}
  \caption{Overview of Learning to Self-Evolve (LSE). \textbf{Left:} Tree-guided self-evolution at test time. Upper Confidence Bound (UCB) selection chooses a context from the evolution tree; the action model generates outputs for a new batch of problems; the self-evolving policy receives the performance summary and proposes a revised context. \textbf{Right:} LSE trains the self-evolving policy via RL with an improvement-based reward computed as the difference between post-edit and pre-edit performance.}
  \label{fig:overview}
\end{figure}

Test-time self-evolution can be characterized along at least two dimensions: \emph{how} the policy is updated and \emph{when}. On one end of the first dimension, gradient-based methods modify model parameters directly; on the other, prompt-based methods rewrite the model context while keeping parameters frozen. Along the second dimension, \emph{intra-episode} evolution updates the policy within a single episode: the model revisits its own attempts and refines its answer to a particular problem, trading additional compute for instance-level gains~\citep{shinn2023reflexion, kumar2025score, yuksekgonul2026learning}. \emph{Inter-episode} evolution updates the policy after one or more completed episodes and applies the result to new problems, extracting transferable knowledge that generalizes across tasks~\citep{yin2024g0del, hu2024adas, zhang2025darwin}.

We focus on inter-episode, prompt-based self-evolution: an LLM observes its performance on a batch of problems and rewrites its own context to improve on the next batch. Several recent works explore this direction through automatic prompt optimization~\citep{khattab2024dspy, agrawal2025gepa, yuksekgonul2024textgrad}, self-referential updates~\citep{fernando2024promptbreeder, zhao2024expel, zhang2025darwin, hu2024adas, zhang2025agentic}, and agentic memory systems~\citep{zhang2025memgen0, zhang2025agentic, chhikara2025mem0}. These methods, however, rely entirely on the inherent ability of the LLM to analyze feedback and propose better context. The model is never explicitly trained for this self-improvement task.

We argue that self-evolution poses a reasoning challenge distinct from other reasoning domains. The process, in essence, shares the structure of an RL problem. An RL optimizer relies on dedicated algorithms to assign credit, estimate gradients, and balance exploration against exploitation. In self-evolution, the model must perform all three implicitly, through natural language reasoning alone. It must judge which parts of the current context help and which hurt, anticipate how a revision will change downstream behavior, and decide whether to refine what works or try something new. These demands motivate explicit optimization for self-evolution.

We propose \textbf{Learning to Self-Evolve (LSE)}, an RL framework that explicitly trains an LLM to be an effective self-evolving policy. Rather than optimizing over the full multi-step evolution trajectory, LSE simplifies training to a single step: the model receives the current context and a performance summary, and produces a better context. Each edit is rewarded by the \emph{improvement} in downstream performance, instead of the absolute post-edit score. At test time, we leverage a tree-guided evolution loop that allows the system to explore and backtrack across possible contexts.

We evaluate LSE on Text-to-SQL generation and general question answering. Despite using only a 4B-parameter model, the LSE-trained policy outperforms self-evolving policies powered by frontier models such as GPT-5 and Claude Sonnet 4.5, as well as prompt optimization methods such as GEPA and TextGrad. Our contributions are as follows:
\begin{itemize}
    \item We formalize test-time inter-episode self-evolution and operationalize it through prompt-based updates with tree-guided search (\S\ref{sec:general_framework}, \S\ref{sec:operationalization}).
    \item We propose LSE, an RL framework that explicitly trains the self-evolving policy with an improvement-based reward (\S\ref{sec:lse}).
    \item We show that a 4B-parameter model trained with LSE outperforms larger untrained models and prompt optimization methods, and transfers to guide other models without additional training (\S\ref{sec:experiments}).
\end{itemize}

\section{Related Work}
\label{sec:related}
The term \emph{self-evolution} has been used to refer to many different concepts in recent LLM research. We organize the landscape into two broad categories. \emph{Training-time self-evolution} focuses on using LLMs to generate their own training data and learning signals during training. \emph{Test-time self-evolution} enables a policy to continue updating itself after training, adapting dynamically based on experience accumulated during deployment.

\paragraph{Training-time self-evolution.}
A growing body of work leverages LLMs to generate their own data and learning signals during training. RL-based post-training has the model produce reasoning traces and optimizes them against verifiable rewards~\citep{lightman2023let, o1, deepseek2025r1}. Bootstrapping methods such as STaR~\citep{zelikman2022star} iteratively generate candidate rationales and fine-tune on the correct ones. Self-rewarding approaches~\citep{yuan2024selfrewarding,zhao2025learning} extend this by using the model itself as the reward signal. Absolute Zero~\citep{zhao2025absolute} takes this to its extreme: a single model both proposes and solves tasks with no external data, using a code executor as the sole source of verifiable reward. While these methods produce stronger models, the resulting policy remains static once training ends. Our work addresses a complementary problem: enabling the policy to continue improving at test time.

\paragraph{Test-time self-evolution.}
A static policy cannot accommodate distribution shifts encountered at test time. Test-time self-evolution addresses this by enabling the model to self-update based on its own experience after deployment. This capability spans two temporal scales. \emph{Intra-episode} methods improve on a single problem instance by allocating additional compute. Reflexion~\citep{shinn2023reflexion} prompts the model to reflect on failed attempts and retry, SCoRe~\citep{kumar2025score} trains self-correction through RL, and TTRL~\citep{zuo2025ttrl} applies RL directly at test time using majority voting as a proxy reward. TTT-Discover~\citep{yuksekgonul2026learning} continues training the model at test time through RL to find the best solution on a single open-ended problem. These methods trade compute for accuracy on individual instances but do not transfer knowledge across problems.

\emph{Inter-episode} methods accumulate experience across completed episodes and apply it to new ones. One active direction is automatic prompt optimization. GEPA~\citep{agrawal2025gepa} and TextGrad~\citep{yuksekgonul2024textgrad} use natural-language feedback from rollouts to iteratively mutate and rewrite prompts. A second direction develops self-referential agents that modify their own code or instructions. ExpeL~\citep{zhao2024expel} extracts transferable lessons from successful and failed trajectories. PromptBreeder~\citep{fernando2024promptbreeder} evolves prompts through mutation and crossover operators. More recent systems such as ADAS~\citep{hu2024adas} and Darwin G\"odel Machine~\citep{zhang2025darwin} extend this by recursively redesigning the self-evolving policy itself~\citep{yin2024g0del}. A third direction builds agentic memory systems: Voyager~\citep{wang2023voyager} accumulates a reusable skill library from experience in Minecraft, while systems such as MemGen~\citep{zhang2025memgen0} and Mem0~\citep{chhikara2025mem0} maintain evolving memory stores that persist across episodes~\citep{zhang2025agentic}. All of these methods rely on the inherent reasoning ability of the LLM to analyze feedback and propose improvements. Our work falls in this category but takes a distinct approach: rather than relying on emergent ability, we explicitly train the self-evolving policy through RL.

\section{Method}
\label{sec:method}

We now introduce our proposed framework and method. We first formalize the test-time inter-episode self-evolution (\S\ref{sec:general_framework}). We then describe how we operationalize it through prompt-based updates and tree-guided search (\S\ref{sec:operationalization}). Finally, we present Learning to Self-Evolve (LSE), an RL framework that trains the self-evolving policy (\S\ref{sec:lse}).

\subsection{Test-Time Inter-Episode Evolution}
\label{sec:general_framework}

Consider a task $\gT = (\gX, \gY, R)$ comprising an input space $\gX$, an output space $\gY$, and a reward function $R: \gX \times \gY \to \R$. A policy $\pi$ maps inputs $x \in \gX$ to outputs $y \in \gY$.
A \emph{self-evolving policy} is a function $f$ that updates the current policy based on experience collected during interaction. Given a task $\gT$, the system executes $T$ rounds of evolution. At each round $t$, the current policy $\pi^{(t)}$ is applied to a batch of $k$ problems sampled from $\gX$, producing experience tuples $\{(x_i, y_i, r_i)\}_{i=1}^k$. The self-evolving policy then computes an updated policy:
\begin{equation}
  \pi^{(t+1)} = f\big(\pi^{(t)},\; \{(x_i, y_i, r_i)\}_{i=1}^k\big).
  \label{eq:general_evolve}
\end{equation}
This produces a sequence of policies $\pi^{(0)}, \pi^{(1)}, \ldots, \pi^{(T)}$. The objective of $f$ is to maximize the cumulative reward over $T$ rounds of evolution:
\begin{equation}
  \sum_{t=0}^{T} \E_{x \sim \gX}\big[R\big(x,\, \pi^{(t)}(x)\big)\big].
  \label{eq:general_obj}
\end{equation}

In the language model setting, a policy $\pi_\theta$ is determined by its parameters $\theta$ and context $c$, comprising system prompts, instructions, skill libraries, and any other textual input that shapes behavior. This decomposition admits two natural instantiations of $f$:
\begin{itemize}
\item \textbf{Gradient-based}: $f$ modifies $\theta$ directly (e.g., via RL or SFT on recent experience);
\item \textbf{Prompt-based}: $f$ modifies $c$ while keeping $\theta$ frozen.
\end{itemize}
We focus on the prompt-based instantiation, where $f$ is itself an LLM that generates updated contexts from past experience. This choice requires no gradient computation at test time, thereby avoiding the catastrophic forgetting problem with continual learning, and casts the evolution problem as a natural-language reasoning task that can itself be improved through training.

\subsection{Prompt-Based Evolution with Tree Search}
\label{sec:operationalization}

An LLM with frozen parameters $\theta$ defines a conditional policy $\pi_\theta(y \mid x, c)$, where $x \in \gX$ is a problem instance and $c$ is the context introduced in \S\ref{sec:general_framework}. In our implementation, we designate a special \emph{instruction field} within $c$ for the self-evolving policy to edit, leaving all other components (e.g., task description, format specification) fixed. 

At each round, the self-evolving policy $f_\psi$ maps the current context and a performance summary to an updated context:
\begin{equation}
  c_{t+1} = f_\psi\big(c_t,\, S_t\big),
  \label{eq:evolve}
\end{equation}
where $S_t = \{(x_i, y_i, y_i^*, r_i)\}_{i=1}^k$ is a \emph{structured performance summary} containing the problems, the outputs of the action LLM, ground-truth answers, and per-problem correctness signals from round $t$.

Note that $S_t$ contains only $k$ problems, where $k$ is typically small, and a different batch is drawn at each round. In-batch performance therefore provides only a noisy estimate of context quality. To obtain a consistent measure across rounds, we fix a separate holdout set $D \subset \gX$ and define the reward of context $c$ as
\begin{equation}
  \bar{R}(c) \;=\; \frac{1}{|D|} \sum_{x \in D} R(x,\, y), \quad y \sim \pi_\theta(\cdot \mid x, c).
  \label{eq:expected_reward}
\end{equation}

\begin{algorithm}[t]
\caption{Prompt-Based Evolution with Tree Search}
\label{alg:self_evolve}
\begin{algorithmic}[1]
\REQUIRE Action policy $\pi_\theta$; self-evolving policy $f_\psi$; task $\gT = (\gX, \gY, R)$; holdout set $D \subset \gX$; initial context $c_0$; rounds $T$; batch size $k$; exploration constant $C$
\STATE Initialize tree $\gG \leftarrow \{(c_0,\, \emptyset,\, \bar{R}(c_0),\, 0)\}$
\FOR{$t = 0, 1, \ldots, T{-}1$}
    \STATE Select node $n^* \leftarrow \argmax_{n \in \gG}\; \bar{R}_n + C\sqrt{(\ln N) / v_n}$ \hfill $\triangleright$ \textit{UCB select}
    \STATE Sample problems $\{x_i\}_{i=1}^k \sim \gX$
    \STATE Generate responses $y_i \sim \pi_\theta(\cdot \mid x_i, c_{n^*})$ for $i = 1, \ldots, k$ \hfill $\triangleright$ \textit{Act}
    \STATE Evaluate $r_i \leftarrow R(x_i, y_i)$ for $i = 1, \ldots, k$ \hfill $\triangleright$ \textit{Evaluate}
    \STATE Construct summary $S_t = \{(x_i, y_i, y_i^*, r_i)\}_{i=1}^k$
    \STATE $c_{\mathrm{new}} \leftarrow f_\psi(c_{n^*}, S_t)$ \hfill $\triangleright$ \textit{Evolve}
    \STATE Evaluate $\bar{R}(c_{\mathrm{new}})$ on holdout set $D$ \hfill $\triangleright$ \textit{Eq.~(\ref{eq:expected_reward})}
    \STATE Append child $(c_{\mathrm{new}},\, S_t,\, \bar{R}(c_{\mathrm{new}}),\, 0)$ to $n^*$ in $\gG$; increment $v_{n^*}$
\ENDFOR
\RETURN $\argmax_{n \in \gG}\; \bar{R}_n$
\end{algorithmic}
\end{algorithm}

\paragraph{Tree-guided evolution.}
The linear evolution chain $c_0 \to c_1 \to \cdots$ greedily extends the most recent context, which risks committing irreversibly to a suboptimal evolution path. To enable broader exploration of the context space, we maintain an \emph{evolution tree} $\gG$ in which each node $n$ stores a tuple $(c_n, S_n, \bar{R}_n, v_n)$ of context, performance summary, mean holdout reward, and visit count. At each round, rather than always extending the latest node, we select the node that maximizes the Upper Confidence Bound (UCB) score~\citep{auer2002using}:
\begin{equation}
  n^* = \argmax_{n \in \gG} \;\bar{R}_n + C\sqrt{\frac{\ln N}{v_n}},
  \label{eq:ucb}
\end{equation}
where $N$ is the number of completed rounds and $C > 0$ controls the exploration-exploitation trade-off. The context and summary of $n^*$ are used as input to the next evolution step, and the resulting child node is appended to $\gG$. This allows the system to revisit and branch from promising contexts discovered earlier, rather than committing to a single evolution path. The full procedure is summarized in Algorithm~\ref{alg:self_evolve}.

\subsection{Learning to Self-Evolve (LSE)}
\label{sec:lse}

While off-the-shelf LLMs already exhibit some ability to iteratively refine their own prompts~\citep{yin2024g0del, agrawal2025gepa, zhang2025darwin}, this ability emerges entirely from pretraining and standard post-training, and the model is never explicitly optimized for self-improvement. We propose Learning to Self-Evolve (LSE), an RL framework that explicitly trains $f_\psi$ to be an effective self-evolving policy.

Recall from Eq.~(\ref{eq:general_obj}) that the goal of the self-evolving policy is to maximize the cumulative reward over 
evolution rounds. A natural training objective for $f_\psi$ is:
\begin{equation}
  \max_{f_\psi}\; \sum_{t=0}^{T} \bar{R}(c_t), \quad \text{where } c_{t+1} = f_\psi(c_t, S_t) \;\;\forall\, t.
  \label{eq:full_obj}
\end{equation}
Directly optimizing this $T$-step objective is costly: each rollout requires $T$ sequential rounds of evaluation and context generation, and the trajectory-level reward introduces a long-horizon credit-assignment problem. We therefore simplify to the \emph{single-step} setting ($T = 1$), where $f_\psi$ produces a single context update $c_1 = f_\psi(c_0, S_0)$ and is rewarded immediately. This reduces the problem to a contextual bandit and avoids the long-horizon credit-assignment difficulty while 
still capturing the core challenge of learning to improve instructions from feedback.

Even in the single-step setting, the choice of reward function is consequential. A natural candidate is the post-edit reward $\bar{R}(c_1)$, the performance of the action policy under the updated context. However, this reward is biased toward contexts that are already effective. Consider two scenarios: (1)~the initial context achieves 80\% accuracy and drops to 70\% after editing, yielding $r = 0.7$; (2)~the initial context achieves 30\% accuracy and improves to 60\%, yielding only $r = 0.6$. The post-edit reward ranks the first scenario higher despite the \emph{degradation} in performance, because it conflates the quality of the starting point with that of the edit. This bias encourages the policy to preserve already-effective contexts rather than genuinely learn to improve them. We instead define the reward as the \emph{improvement in reward}:
\begin{equation}
  r_{\mathrm{LSE}} \;=\; \bar{R}(c_1) - \bar{R}(c_0),
  \label{eq:meta_reward}
\end{equation}
which directly incentivizes $f_\psi$ to produce edits that improve performance relative to the starting point, regardless of the initial performance level.

Notably, if $r_{\mathrm{LSE}}$ is used as the reward in a standard policy-gradient algorithm such as PPO or GRPO, the baseline estimator absorbs the $\bar{R}(c_0)$ term. To see this, let $s = (c_0, S_0)$ denote the state observed before the edit. A baseline $V(s)$ estimated under $r_{\mathrm{LSE}}$ satisfies $V'(s) = \E[\bar{R}(c_1) - \bar{R}(c_0) \mid s] = V(s) - \bar{R}(c_0)$, where $V(s) = \E[\bar{R}(c_1) \mid s]$ is the baseline under the post-edit reward. The advantage then reduces to:
\begin{equation}
  A'(s, c_1) \;=\; r_{\mathrm{LSE}} - V'(s) \;=\; \big(\bar{R}(c_1) - \bar{R}(c_0)\big) - \big(V(s) - \bar{R}(c_0)\big) \;=\; \bar{R}(c_1) - V(s),
\end{equation}
which is identical to the advantage under the post-edit reward alone. That is, the delta-reward and the post-edit reward yield 
the same gradient estimates whenever a learned baseline is used. Rather than using a value model or group-based normalization 
as the baseline, we can bypass baseline estimation entirely: the pre-edit reward $\bar{R}(c_0)$ is known before $f_\psi$ acts and equals the reward of a null edit that returns $c_0$ unchanged, so it can serve directly as the baseline. This yields the LSE advantage:
\begin{equation}
  A_{\mathrm{LSE}} \;=\; \bar{R}(c_1) - \bar{R}(c_0),
  \label{eq:lse_advantage}
\end{equation}
and the corresponding policy-gradient estimate:
\begin{equation}
  \nabla_\psi J \;=\; \E_{c_1 \sim f_\psi(\cdot \mid c_0, S_0)} \Big[ A_{\mathrm{LSE}} \;\nabla_\psi \log f_\psi(c_1 \mid c_0, S_0) \Big].
  \label{eq:lse_gradient}
\end{equation}
Because $\bar{R}(c_0)$ is action-independent, using it as a baseline does not alter the expected gradient. It is, however, a control variate that cancels prompt-specific offsets. In practice, evaluation noise and between-prompt difficulty variation likely dominate raw accuracy scores. Under these conditions, the improvement-based advantage provides a cleaner learning signal and more stable policy-gradient updates. It also reduces training cost, as it requires neither multiple rollouts per prompt for group-based normalization nor a separate value network.

The gradient in Eq.~(\ref{eq:lse_gradient}) depends on the distribution of starting states $s = (c_0, S_0)$. If $c_0$ is always the seed context, a mismatch arises: at test time, the policy runs for multiple rounds (Algorithm~\ref{alg:self_evolve}) and must improve contexts produced by its own prior edits. We therefore populate the tree $\gG$ with multiple rounds of evolution to construct the training dataset, then randomly sample nodes from $\gG$ as starting contexts at every RL step. This exposes $f_\psi$ to a distribution of contexts similar to what it will see during multi-step evolution.

\section{Experiments}
\label{sec:experiments}

We evaluate LSE on two task domains, Text-to-SQL generation and general question answering, comparing against both stronger models and alternative prompt optimization methods (\S\ref{sec:setup}--\ref{sec:main_results}). We then ablate the reward design and search strategy in \S\ref{sec:analysis}.

\subsection{Experimental Setup}
\label{sec:setup}

\begin{table}[t]
\caption{Text-to-SQL results on BIRD. All methods use Qwen3-4B-Instruct as the action policy~$\pi_\theta$. We report execution accuracy~(\%). Best result per column in \textbf{bold}.}
\label{tab:bird}
\centering
\small
\begin{tabular}{l ccccc c}
\toprule
Method & Financial & Toxicology & Codebase & Formula~1 & Card Games & Avg. \\
\midrule
Seed prompt & 51.0 & 60.3 & 63.7 & 54.5 & 56.5 & 57.2 \\
Qwen3-4B-Instruct & 63.7 & 60.3 & 70.2 & 56.0 & 61.0 & 62.2 \\
\midrule
Claude Sonnet 4.5 & 70.8 & 63.8 & 67.8 & 57.3 & 63.0 & 64.5 \\
GPT-5 & 70.8 & 65.8 & \textbf{72.0} & 54.3 & 63.3 & 65.2 \\
\midrule
GEPA & 64.0 & 62.0 & \textbf{72.0} & 54.0 & 62.0 & 62.8 \\
TextGrad & 60.3 & 66.0 & 71.5 & 56.5 & 61.3 & 63.1 \\
\midrule
LSE (ours) & \textbf{72.0} & \textbf{68.5} & \textbf{72.0} & \textbf{59.8} & \textbf{64.0} & \textbf{67.3} \\
\bottomrule
\end{tabular}
\end{table}

\paragraph{Models.}
We use Qwen3-4B-Instruct as both the action policy $\pi_\theta$ and the self-evolving policy $f_\psi$, unless otherwise specified. Training details and hyperparameters can be found in Appendix~\ref{app:training}.

\paragraph{Tasks and datasets.}
We evaluate on tasks across two domains: Text-to-SQL generation, where the policy produces executable SQL queries that retrieve the data specified by a user question, and general question answering (QA), where the policy answers multiple-choice questions across diverse academic subjects. The prompts for each task are in Appendix~\ref{app:prompts}.

For Text-to-SQL, we use BIRD~\citep{li2024can}, which pairs natural-language questions with SQL queries across database domains. Each database is a separate task domain: problems are sampled from the same domain for both evolution rounds and holdout evaluation. We train on the BIRD training split and evaluate on five randomly selected databases from the BIRD-SQL Mini-Dev split.

For general QA, we use SuperGPQA~\citep{pteam2025supergpqascalingllmevaluation} and MMLU-Redux~\citep{gema2024mmlu}. We convert SuperGPQA questions to four-way multiple-choice format to match MMLU-Redux, and treat each subject as a separate task domain. As with Text-to-SQL, each evolution run operates within a single subject domain. We train on SuperGPQA and evaluate on ten subjects from MMLU-Redux.

\paragraph{Baselines.}
We first evaluate stronger models as the self-evolving policy $f_\psi$ while keeping Qwen3-4B-Instruct as the action policy $\pi_\theta$. We consider two frontier closed-source models, GPT-5~\citep{singh2025openai} and Claude Sonnet 4.5~\citep{sonnet45}

We also compare with two alternative designs of the self-evolving policy. For both methods, we use Qwen3-4B-Instruct as the prompt proposer and optimizer.
\begin{itemize}
    \item \textbf{GEPA}~\citep{agrawal2025gepa} is a reflective prompt optimizer that merges textual reflection with multi-objective evolutionary search. GEPA mutates prompts based on natural-language feedback from new rollouts and maintains a Pareto front over per-instance performance to avoid greedy local optima. Each GEPA optimization step corresponds to one evolution round: the sampled problem batch is the training data for reflection, and the holdout set $D$ is $D_{\mathrm{pareto}}$ in GEPA. 
    \item \textbf{TextGrad}~\citep{yuksekgonul2024textgrad} decomposes each prompt update into two LLM calls: a \emph{backward} call that critiques the current instruction given the batch failures and produces natural-language ``gradients'' (feedback on how the instruction should change), followed by a \emph{Textual Gradient Descent (TGD)} call that rewrites the instruction by incorporating the feedback. We follow the example provided in the official repository\footnote{\url{https://github.com/zou-group/textgrad/blob/main/evaluation/prompt_optimization.py}} and treat each backward--TGD step as one evolution round.
\end{itemize}

\paragraph{Evaluation protocol.}
For all methods, we sample problems and present them to the action policy in a fixed order across runs. The holdout set $D$ is also fixed across all evaluation runs within each task domain. We report the best performance achieved over $T$ rounds of evolution. Additional details can be found in Appendix~\ref{app:training}.

\subsection{Main Results}
\label{sec:main_results}

\begin{table}[t]
\caption{Question-answering results on MMLU-Redux. All methods use Qwen3-4B-Instruct as the action policy~$\pi_\theta$. We report accuracy~(\%). Best result per column in \textbf{bold}.}
\label{tab:mmlu}
\centering
\resizebox{\textwidth}{!}{
\begin{tabular}{l cccccccccc c}
\toprule
Method & Bus.~Eth. & Phil. & Prof.~Acc. & Econ. & Anat. & Security & Virol. & Moral~Sc. & Glob.~Facts & Prof.~Law & Avg. \\
\midrule
Seed prompt & 70.5 & 76.0 & 67.7 & 76.2 & 76.0 & 68.2 & 72.5 & 59.7 & 55.5 & 54.0 & 67.6 \\
Qwen3-4B-Instruct & 75.5 & 79.2 & 73.0 & 83.0 & 82.3 & 68.8 & 77.0 & 59.7 & 56.0 & 57.3 & 71.2 \\
\midrule
Claude Sonnet 4.5 & 75.3 & 79.0 & 75.5 & 81.0 & 79.8 & 69.3 & 77.5 & \textbf{67.3} & \textbf{57.0} & 58.5 & 72.0 \\
GPT-5 & \textbf{77.5} & 80.0 & 73.0 & 83.5 & 81.0 & 70.0 & \textbf{82.0} & 64.8 & 55.5 & 58.0 & 72.5 \\
\midrule
GEPA & 76.0 & 80.0 & \textbf{78.0} & 84.0 & 82.0 & 70.0 & 78.0 & 62.0 & 56.0 & \textbf{64.0} & 73.0 \\
TextGrad & 74.0 & 79.0 & 67.8 & 73.3 & 80.0 & 67.8 & 77.3 & 62.5 & 51.3 & 58.3 & 69.1 \\
\midrule
LSE (ours) & 75.0 & \textbf{82.0} & 73.0 & \textbf{85.0} & \textbf{84.5} & \textbf{70.5} & 79.0 & 64.5 & \textbf{57.0} & 62.0 & \textbf{73.3} \\
\bottomrule
\end{tabular}
}
\end{table}

Tables~\ref{tab:bird} and~\ref{tab:mmlu} report results on Text-to-SQL and QA. Even without explicit training for self-improvement, off-the-shelf LLMs can refine their own prompts when given test-time feedback. The untrained Qwen3-4B-Instruct baseline improves over the seed prompt by 5\% on BIRD and 3.6\% on MMLU-Redux. This confirms that LLMs can already learn from their own experience within the evolution loop of Algorithm~\ref{alg:self_evolve}.

RL training with LSE substantially improves this ability. Despite using only a 4B-parameter model, LSE outperforms both frontier models on BIRD, surpassing GPT-5 by 2.1\% (67.3\% vs.\ 65.2\%) and Claude Sonnet 4.5 by 2.8\%. On MMLU-Redux, LSE matches GPT-5 (73.3\% vs.\ 72.5\%) and outperforms Claude Sonnet 4.5. These results indicate that explicit RL training for self-evolution is effective, enabling a small model to match or surpass frontier models.

LSE also outperforms both prompt optimization methods. On BIRD, LSE surpasses GEPA by 4.5\% (67.3\% vs.\ 62.8\%) and TextGrad by 4.2\% (67.3\% vs.\ 63.1\%). On MMLU-Redux, LSE matches GEPA (73.3\% vs.\ 73.0\%) and outperforms TextGrad by 4.2\%. Together, these results show that while off-the-shelf LLMs have some prompt-refinement ability, explicit training to self-evolve matches or outperforms untrained baselines, larger models, and specialized optimization methods.

Finally, both the improvement from self-evolution over the seed prompt and the additional benefit of LSE are smaller on MMLU-Redux than on BIRD. One possible explanation is the structure of the two tasks. In BIRD, all queries within a domain target the same database, so there is clear shared knowledge across problems: understanding the schema, common join patterns, or column semantics from one query directly helps with others. In MMLU-Redux, problems within the same subject are deliberately deduplicated and designed to cover broad topics. Solving one econometrics question does not guarantee useful knowledge for the next. This limits how much any self-evolving policy can improve the action policy's context from experience within a single domain.

\subsection{Analysis}
\label{sec:analysis}

\begin{figure}[t]
  \centering
  \begin{subfigure}[b]{0.48\linewidth}
    \centering
    \includegraphics[width=\linewidth]{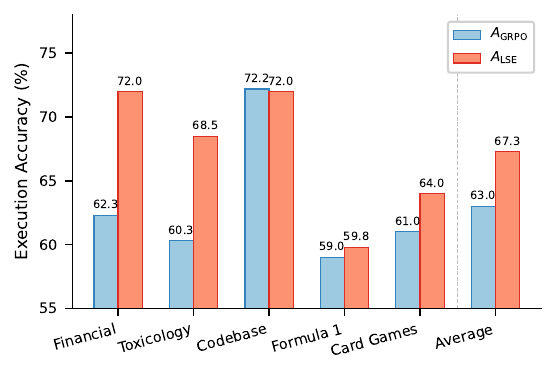}
    \caption{Effect of reward design}
    \label{fig:ablation_reward}
  \end{subfigure}
  \hfill
  \begin{subfigure}[b]{0.48\linewidth}
    \centering
    \includegraphics[width=\linewidth]{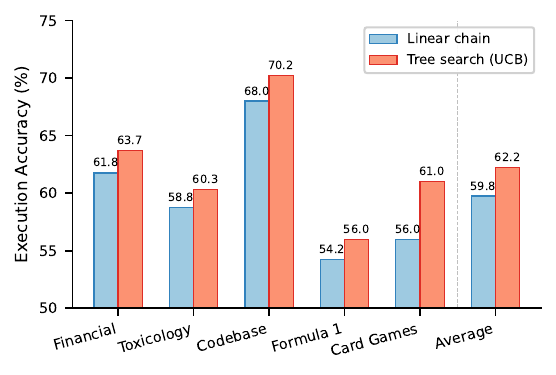}
    \caption{Effect of search strategy}
    \label{fig:ablation_search}
  \end{subfigure}
  \caption{Ablation studies on reward design and search strategy. (a)~$A_{\mathrm{GRPO}}$ uses $\bar{R}(c_1)$ with GRPO's group-based advantage; $A_{\mathrm{LSE}}$ uses the improvement-based reward $r_{\mathrm{LSE}} = \bar{R}(c_1) - \bar{R}(c_0)$ (Eq.~\ref{eq:meta_reward}). (b)~Tree search (UCB) vs.\ linear chain (always extends the most recent node), both with the untrained Qwen3-4B-Instruct as~$f_\psi$.}
  \label{fig:ablations}
\end{figure}

\paragraph{Effect of reward design.}
In \S\ref{sec:lse} we motivated $A_{\mathrm{LSE}} = \bar{R}(c_1) - \bar{R}(c_0)$ as a cleaner learning signal than the standard GRPO advantage $A_{\mathrm{GRPO}}$, which reduces to optimizing post-edit accuracy. We train a variant with $A_{\mathrm{GRPO}}$, keeping all other settings identical. On BIRD, $A_{\mathrm{LSE}}$ outperforms $A_{\mathrm{GRPO}}$ by 4.3\% (67.3\% vs.\ 63.0\%; Figure~\ref{fig:ablation_reward}). These results provide empirical evidence that the improvement-based objective is more effective for training self-evolving policies.

\paragraph{Effect of search strategy.}
We compare UCB tree search against a linear-chain baseline that always extends the most recent node. Both use the untrained Qwen3-4B-Instruct as $f_\psi$. Figure~\ref{fig:ablation_search} shows that tree search improves the average by 2.4\% on BIRD (62.2\% vs.\ 59.8\%) and 2.2\% on MMLU-Redux (71.2\% vs.\ 69.0\%; Figure~\ref{fig:search_mmlu}).

\begin{wrapfigure}{r}{0.48\linewidth}
  \vspace{-20pt}
  \centering
  \includegraphics[width=\linewidth]{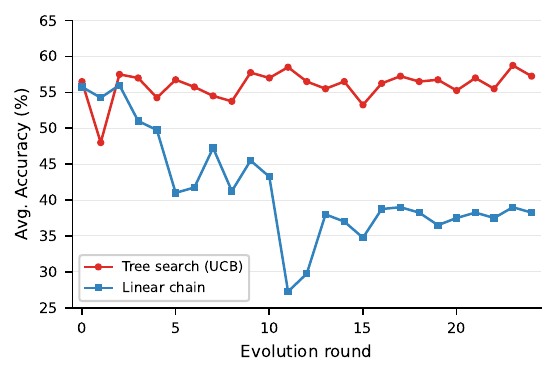}
  \caption{Per-round average accuracy on the BIRD Card Games database. The linear chain cannot recover from bad edits, while tree search (UCB) backtracks to higher-scoring ancestors.}
  \label{fig:search_trajectory}
  \vspace{-30pt}
\end{wrapfigure}

The key advantage is that tree search does not commit to a bad edit irrevocably. Figure~\ref{fig:search_trajectory} shows a concrete example on the BIRD Card Games split. The linear chain's average accuracy collapses from 56\% to below 30\% after a sequence of bad edits, and never recovers because each round builds on the previous context. With tree search, a bad edit at an early round does not cascade: UCB selection shifts back to a higher-scoring ancestor, keeping the trajectory out of bad local optima.

\paragraph{Test-time self-evolution for specialized models.}
Current LLM development often involves training specialized models for a domain of tasks. Can test-time self-evolution further improve such models? We test this by replacing $\pi_\theta$ with Arctic-Text2SQL-R1-7B~\citep{yao2025arctic}, a text-to-SQL model fine-tuned with RL on the BIRD training set, and applying the same LSE-trained $f_\psi$ (Qwen3-4B-Instruct) without additional training.

Table~\ref{tab:transfer} shows that LSE evolution improves Arctic by 6.7\% on average (57.7\% $\to$ 64.4\%). This indicates that parameter-level and prompt-level optimization are complementary: RL training encodes general SQL patterns into model weights, while prompt evolution adapts the context to each database at test time. The result also demonstrates that the LSE-trained policy transfers across action models: although $f_\psi$ was trained exclusively with Qwen3-4B-Instruct, the evolution strategy generalizes to guide a different model.

\begin{table}[t]
\caption{Test-time self-evolution of a specialized action model on BIRD. Arctic-Text2SQL-R1-7B~\citep{yao2025arctic}, an RL-tuned text-to-SQL model, serves as the action policy~$\pi_\theta$. The self-evolving policy~$f_\psi$ is the LSE-trained Qwen3-4B-Instruct from the main experiments, applied without further training. We report execution accuracy~(\%).}
\label{tab:transfer}
\centering
\small
\begin{tabular}{l ccccc c}
\toprule
Variant & Financial & Toxicology & Codebase & Formula~1 & Card Games & Avg. \\
\midrule
Seed prompt & 56.8 & 54.5 & 65.3 & 52.3 & 59.5 & 57.7 \\
+ LSE evolution & 68.3 & 62.3 & 71.5 & 57.0 & 63.0 & 64.4 \\
\bottomrule
\end{tabular}
\end{table}

\section{Conclusion}

This work demonstrates that test-time self-evolution is a learnable skill that can be directly optimized through fine-tuning. The central design choice in LSE is a single-step RL objective that rewards the improvement each edit produces, sidestepping multi-step trajectory optimization while still capturing the core challenge of learning from feedback. Tree-guided search then composes these edits into multi-round evolution at test time. Our results show that direct optimization for self-evolution is effective, enabling a 4B-parameter model to match or surpass frontier models and prompt optimizers. Taken together, these findings highlight the benefit of targeting self-evolution as a distinct skill and designing learning algorithms for it.

\paragraph{Limitations.} Our work has several limitations. First, we reduce the multi-step evolution problem to a single-step training objective, delegating exploration entirely to the tree search algorithm at test time. Jointly optimizing over multi-step trajectories could yield stronger policies but would introduce additional challenges in credit assignment and computational cost. Second, we train a separate self-evolving policy for each task domain. Training a single policy that generalizes across diverse tasks is a natural extension, though it likely requires large-scale training across many domains. Third, we constrain evolution to the instruction field of the context; other components such as tools, skill libraries, and external memory are not explored. More broadly, the LSE framework could be paired with updates in the latent space or parameter space~\citep{sun19ttt, tandon2025end0to0end}. Finally, our training and evaluation environments are relatively small in scale. Curating effective environments for test-time self-evolution is difficult, as it requires not only sufficient problems with feedback but also problems that share enough structure for evolution to be meaningful. Developing more principled and scalable approaches to environment curation and evaluation remains an important open problem.

\bibliography{iclr2026_conference}
\bibliographystyle{iclr2026_conference}

\appendix

\section{Training and Evaluation Details}
\label{app:training}

\paragraph{Data generation.}
We train a separate self-evolving policy for each task domain (Text-to-SQL and QA) using evolutions trajectories generated from the corresponding training set. For each domain, we run 200 data-generation runs, each containing 20 rounds of evolution, yielding approximately 4{,}000 tree nodes to sample from during RL training.

\paragraph{RL training.}
We implement our RL framework using verl~\citep{sheng2024hybridflow}. We find that randomly sampling nodes from the evolution trees produces weak training signal early in training. Instead, we build a simple curriculum by preferentially sampling nodes with the highest improvement potential, defined as the difference between a node's performance and the maximum performance in its own tree. We use a learning rate of $1 \times 10^{-5}$, sample 32 nodes per batch, and generate 4 rollouts per node. We perform on-policy training and do not apply KL regularization. We train for 4 epochs and select the best checkpoint based on a separate development set.

\paragraph{Evaluation protocol.}
For every domain, we fix the holdout set $D$ at 50 problems. Performance on the holdout set is calculated as the average over eight generations. We run 25 rounds of evolution and report the best holdout performance achieved by each self-evolving method over the course of evolution. At each round, a batch of 10 problems is sampled with replacement and presented to the action model. The random seed is fixed so that all methods observe the same sequence of problem batches.

\paragraph{Dataset sizes.}
Table~\ref{tab:dataset_sizes} reports the number of evaluation problems per domain.

\begin{table}[h]
\centering
\caption{Number of evaluation problems per domain.}
\label{tab:dataset_sizes}
\small
\begin{tabular}{ll r}
\toprule
Task & Domain & \# Problems \\
\midrule
\multirow{5}{*}{BIRD} & Financial & 106 \\
 & Toxicology & 145 \\
 & Card Games & 191 \\
 & Formula 1 & 174 \\
 & Codebase & 186 \\
\midrule
\multirow{10}{*}{MMLU-Redux} & Business Ethics & 100 \\
 & Philosophy & 100 \\
 & Professional Accounting & 95 \\
 & Econometrics & 100 \\
 & Anatomy & 100 \\
 & Security Studies & 100 \\
 & Virology & 96 \\
 & Moral Scenarios & 100 \\
 & Global Facts & 96 \\
 & Professional Law & 100 \\
\bottomrule
\end{tabular}
\end{table}

\begin{figure}[h]
  \centering
  \includegraphics[width=\linewidth]{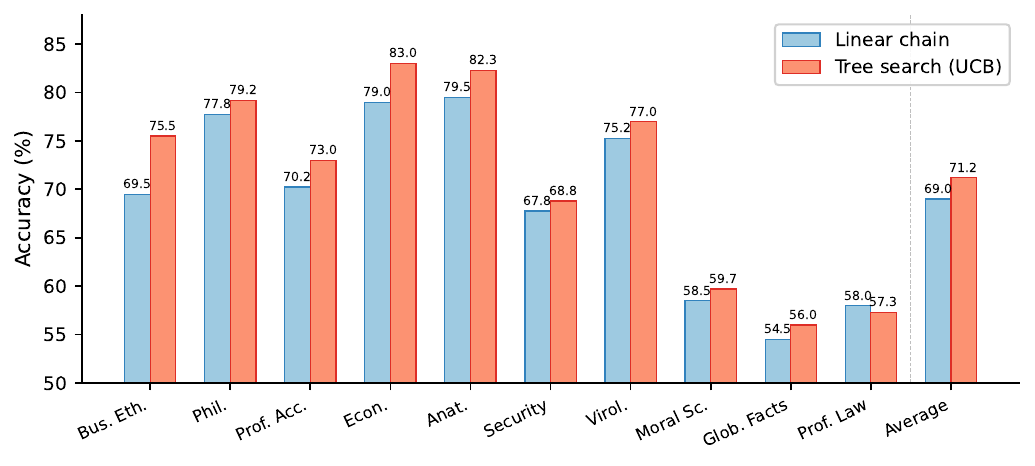}
  \caption{Search strategy ablation on MMLU-Redux, complementing Figure~\ref{fig:ablation_search}. Both variants use the untrained Qwen3-4B-Instruct as the self-evolving policy~$f_\psi$. Tree search improves the average accuracy from 69.0\% to 71.2\%.}
  \label{fig:search_mmlu}
\end{figure}

\section{Prompts}
\label{app:prompts}

Each task uses three prompt templates: (1)~a \emph{system prompt} that provides the action model $\pi_\theta$ with task context and the current instructions, (2)~a \emph{user message} that presents each problem instance, and (3)~a \emph{self-evolution prompt} that the self-evolving policy $f_\psi$ receives to produce a revised instruction. The instruction field within the system prompt is the component that $f_\psi$ edits at each evolution round. Below we reproduce the templates for both tasks.

\subsection{Text-to-SQL}

\begin{framed}
\noindent\textbf{Action model system prompt.}
\begin{small}
\begin{verbatim}
Task Overview:
You are a data science expert. Below, you are provided with a
database schema and a natural language question. Your task is to
understand the schema and generate a valid SQL query to answer the
question.

Database Engine:
SQLite

Database Schema:
{schema}
This schema describes the database's structure, including tables,
columns, primary keys, foreign keys, and any relevant relationships
or constraints.

**Instructions**
{instructions}
\end{verbatim}
\end{small}
The seed instruction is: \texttt{Return only a single valid SQLite SQL statement in <answer>...</answer>.}
\end{framed}

\begin{framed}
\noindent\textbf{Action model user message.}
\begin{small}
\begin{verbatim}
Question:
{question}

**Instructions**
{instructions}

Follow the instructions and show your work. When you are ready,
return the query output list in tags: <answer> ... </answer>
\end{verbatim}
\end{small}
\end{framed}

\begin{framed}
\noindent\textbf{Self-evolving policy prompt.}
\begin{small}
\begin{verbatim}
You are an expert at designing text-to-SQL agents. The agent is
running on a fixed database schema. Below is the current agent
prompt and a summary of recent performance. Rewrite ONLY the
instructions to improve execution accuracy while maintaining
strict output format.

Current prompt:
{old_prompt}

Evaluation summary over {n_problems} problems and the agent's
full thinking process:
{summary}

**How to write Instructions**
- The agent will continue receive different user queries so don't
  make the instructions too specific to a single question.
  Referring to the questions in the current summary with only the
  question number is not helpful.
- Keep it concise and practical.
- You may include rules, heuristics, knowledge about the database,
  low-level instructions/examples, high-level ideas/strategies,
  pitfalls and any information that you think can make the agent
  better.
- Organize however you like (bullets, headings, checklists).
- Be creative and think about the agent's behavior across
  iterations. Don't be confined by what I told you.
- Don't change the output format, the agent should still return
  the final SQL query in tags: <answer> ... </answer>.

Think step by step and show your work. Reason about the history
of the model's behavior across iterations.

When you are ready, put your revised Instructions within
<prompt>[your new instructions]</prompt> tags.
\end{verbatim}
\end{small}
\end{framed}

\subsection{Question Answering}

\begin{framed}
\noindent\textbf{Action model system prompt.}
\begin{small}
\begin{verbatim}
Task Overview:
You are an expert taking a test. Below, you are provided with a
question and a list of choices. Your task is to select the correct
answer from the choices.

**Instructions**
{instructions}
\end{verbatim}
\end{small}
The seed instruction is: \texttt{Return only the letter of the correct choice (A, B, C, or D) in <answer>...</answer>.}
\end{framed}

\begin{framed}
\noindent\textbf{Action model user message.}
\begin{small}
\begin{verbatim}
Question:
{question}

Choices:
{choices}

Follow the instructions and show your work. When you are ready,
return the answer letter in tags: <answer> ... </answer>
\end{verbatim}
\end{small}
\end{framed}

\begin{framed}
\noindent\textbf{Self-evolving policy prompt.}
\begin{small}
\begin{verbatim}
You are an expert at designing agents for solving multiple-choice
questions that involve both factual knowledge and reasoning.
Below is the current agent prompt and a summary of recent
performance on a set of problems. Rewrite ONLY the instructions
to improve accuracy while maintaining strict output format.

Current prompt:
{old_prompt}

Evaluation summary over {n_problems} problems and the agent's
full thinking process:
{summary}

**How to write Instructions**
- The agent will continue to receive different questions from the
  same subjects. Don't make the instructions too specific to a
  single question.
- Keep it concise and practical.
- You may include rules, heuristics, strategies for multiple
  choice questions (e.g., elimination, careful reading), knowledge
  about the subjects (e.g., common misconceptions, important
  facts, etc.), and any information that you think can make the
  agent better.
- Organize however you like (bullets, headings, checklists).
- Be creative and think about the agent's behavior across
  iterations. Don't be confined by what I told you.
- Don't change the output format, the agent should still return
  the final answer letter in tags: <answer> ... </answer>.

Think step by step and show your work. Reason about the history
of the model's behavior across iterations.

When you are ready, put your revised Instructions within
<prompt>[your new instructions]</prompt> tags.
\end{verbatim}
\end{small}
\end{framed}

\subsection{Example Instructions Found by LSE}

Below we reproduce the best instructions discovered by the LSE-trained self-evolving policy on one domain from each task. These instructions replace the seed instruction in the action model prompt after multiple rounds of tree-guided evolution.

\begin{framed}
\noindent\textbf{Text-to-SQL: BIRD Codebase database.}
\begin{small}
\begin{verbatim}
1. Always return exactly one valid SQLite query in the format:
   <answer>YOUR_SQL_QUERY_HERE</answer>

2. Carefully analyze the natural language question to identify:
   - The target attribute (e.g., name, count, date, status)
   - The relevant tables and their columns
   - Any joins required to connect tables via foreign keys
   - Any filters (e.g., equality, date ranges, null checks)

3. Use only the provided schema. Do not assume columns or
   tables that are not defined.

4. When joining tables:
   - Match foreign key references precisely
     (e.g., posts.OwnerUserId -> users.Id)
   - Use explicit column aliases only if needed for clarity
   - Ensure join conditions match the schema

5. For date filtering:
   - Use strftime('%Y', column) to extract year
   - Never use YEAR() -- SQLite does not support it
   - Match date format exactly (e.g., '2014-04-23 20:29:39.0')

6. For conditional outputs (e.g., "well-finished"):
   - Use CASE WHEN or IIF to map NULL / non-NULL values
   - Match the definition in the question
     (e.g., "not well-finished" = ClosedDate IS NULL)

7. Common pitfalls to avoid:
   - Misidentifying OwnerUserId vs. LastEditorUserId
   - Incorrectly joining on UserId instead of Id
   - Misspelling column names (e.g., CreaionDate)
   - Forgetting required joins for user attributes
   - Confusing UserDisplayName in comments with post ownership

8. Always use subqueries for exact values (MIN, MAX):
   - e.g., WHERE Age = (SELECT MIN(Age) FROM users)
     instead of ORDER BY Age LIMIT 1

9. For percentages or ratios, compute numerator and
   denominator separately using subqueries. Use
   CAST(... AS REAL) for floating-point division.

10. Avoid redundant joins -- if a query can be answered
    from a single table, do not introduce unnecessary joins.
\end{verbatim}
\end{small}
\end{framed}

\begin{framed}
\noindent\textbf{Question answering: MMLU-Redux Anatomy subject.}
\begin{small}
\begin{verbatim}
- Return only the letter of the correct choice (A, B, C,
  or D) in <answer>...</answer>.
- Carefully analyze the question and all answer options
  before selecting.
- Use elimination to rule out clearly incorrect choices
  based on factual knowledge or logical inconsistency.
- For biological and anatomical questions, recall key
  structures and their functions (e.g., fertilization
  occurs in the fallopian tube, not the ovary or uterus;
  the pituitary is the "master gland").
- In neurology: upper motor neuron lesions cause spastic
  paralysis; lower motor neuron lesions cause flaccid
  paralysis; sympathetic pathways use noradrenaline.
- In embryology, palatine shelf elevation is due to turgor
  pressure from hydrophilic molecules, not directly from
  tongue descent or brain flexure.
- For spinal cord injuries, breathing is controlled by the
  brainstem (medulla), not the cervical spinal cord.
- Fracture types: closed = skin intact; greenstick = bent
  but not displaced; open = skin broken; spiral = twisting.
- In autonomic responses, sympathetic chain damage leads to
  pupillary constriction and vasodilation (loss of
  vasoconstriction), not increased sweating.
- In Horner's syndrome: miosis, facial vasodilation,
  decreased lacrimation, and anhydrosis.
- In cerebrovascular accidents, internal capsule lesions
  cause contralateral spastic paralysis.
- Prioritize accuracy over common assumptions -- especially
  regarding directionality (contralateral vs ipsilateral),
  timing (diastole vs systole), and structural relationships.
- Be vigilant for common misconceptions used as distractors.
- Do not over-rely on common associations; base decisions
  on precise anatomical, physiological, or pathological facts.
\end{verbatim}
\end{small}
\end{framed}

\end{document}